\documentclass{article}
\PassOptionsToPackage{numbers,sort&compress}{natbib}

\usepackage[dblblindworkshop,final]{neurips_2026}

\workshoptitle{Agentic Intelligence for Medical (AIM-NeurIPS) Imaging and Multimodal Clinical Data}
\usepackage[utf8]{inputenc}
\usepackage[T1]{fontenc}
\usepackage{booktabs}
\usepackage{microtype}
\usepackage{amsmath}
\usepackage{xcolor}
\usepackage{url}
\usepackage{graphicx}

\title{Label Agreement Does Not Measure Authorization}

\author{
  Amir Sabbaghziarani$^{1}$ \quad
  Bradley Thomas Baker$^{1}$ \quad
  Theodore J. LaGrow$^{2}$ \quad
  Sergey Plis$^{1}$ \\[2mm]
  $^{1}$Joint Georgia State University / Georgia Institute of Technology / Emory University\\
  Center for Translational Research in Data Science and Neuroimaging \\
  $^{2}$Georgia Institute of Technology
}

\begin{document}
\maketitle

\begin{abstract}
Many groups now delegate label ontology and metadata harmonization to agentic LLM pipelines. We built one and audited it. Our aggregate scores looked healthy, but the pipeline kept failing in ways they did not show, so we set out to find what they hid. Label agreement asks whether a proposed label matches a reference. It does not ask whether the agent was entitled to propose it, whether the output was complete enough to act on, or whether the label moved when the evidence moved. We measured those three separately on COBRE and FBIRN, two schizophrenia and control neuroimaging cohorts from different consortia, and they come apart, from label agreement and from each other. Showing the agent an upstream proposal barely moves label agreement, 0.857 to 0.870, while agreement on the chosen action doubles, 0.409 to 0.830. Output that parses as JSON still drops a required field on 10\% of one model's cases and 33\% of the other's. And an agent that replays its first answer scores perfectly on original cases and zero once we change the evidence that decides them. Downstream, a row-order error that none of these metrics reports erases most of the diagnostic signal. So we measure these properties apart, pair each with a control, and gate commitment on the result, which makes failures visible and easy to route to a person. None of this prevents failure. Our reference labels are rule-derived, so agreement with them means consistency, not correctness. Code is available at \url{https://github.com/amir-sbg/Label-Agreement-Does-Not-Measure-Authorization}.
\end{abstract}

\section{Introduction}

Many groups now use LLMs to harmonize label ontologies and metadata across medical datasets. By \emph{harmonization} we mean mapping source files and variables onto a common representation while preserving source meaning, coding, units, and identity relations, not ComBat-style removal of site and scanner effects \citep{johnson2007combat,fortin2017combat}. It is hard in neuroimaging repositories because the justifying evidence is scattered across clinical tables, dictionaries, identifier files, MATLAB bridges, and imaging derivatives, and repositories differ in structure, naming, and identifiers \citep{aine2017cobre,keator2016fbirn,keator2013nidm}.

We work with COBRE and FBIRN, two neuroimaging cohorts of people with schizophrenia and healthy controls that different consortia collected and curated, so the same clinical question reaches us in two encodings. Their phenotype tables carry short, undocumented column names whose meaning must be inferred from surrounding evidence: \texttt{Dx} holds diagnosis codes with no record of whether $0$ or $1$ denotes schizophrenia, \texttt{CPZ} holds chlorpromazine-equivalent doses with no verified unit, and \texttt{SubjectID}, \texttt{SiteID}, \texttt{VisitID} are joinable while denoting different entities. In each case a model can emit a plausible label with an action the evidence does not support, and one accuracy number cannot tell the two apart.

\textbf{Related work.} LLM schema matching is an active area, and controlled studies show added context can hurt \citep{parciak2024schema,korini2023column}. Biomedical systems add ontology constraints, interactive validation, and harmonization toolkits \citep{hardi2026standardization,wu2026bdiviz,lopez2026bdikit}, and LLM workflows address general schema integration \citep{korini2026sintflow}; our residual question is what authorizes a mapping when parser, value, lineage, or imaging-order evidence is missing. Two findings shape the design: brain-MRI VLM audits report poor verbal-confidence calibration \citep{sabbaghziarani2026confident}, and MetaHarmonizer shows public identifiers can inflate apparent performance \citep{li2026metaharmonizer}. Validation-metric analyses show one score can hide distinct failure modes \citep{maierhein2024metrics}; our controlled-error test applies that to an agent as a contrast set \citep{gardner2020contrast,ribeiro2020checklist}, complementing agent-interaction and safety benchmarks \citep{yao2024taubench,ruan2024toolemu}. Evidence Cards build on provenance and documentation practice \citep{w3cprov,gorgolewski2016bids,keator2013nidm,sahoo2023neurobridge,gebru2021datasheets,pushkarna2022datacards}; abstention is a valid selective action \citep{geifman2017selective,wen2025abstention}.

We built such a pipeline and audited it. Its aggregate scores looked healthy while it kept failing, so we asked what those scores were not asking. Label agreement asks whether a proposed label matches a reference. It does not ask whether the agent was entitled to propose it (Section~\ref{sec:auth}), whether the output was complete enough to act on (Section~\ref{sec:contract}), or whether the label moved when the evidence moved (Section~\ref{sec:cec}). We contribute a bounded pipeline separating evidence extraction, interpretation, and authorization; a diagnostic for each of those three questions and for downstream consequence; and an audit showing they come apart, from label agreement and from each other.


\section{Methods}

\begin{figure}[t]
    \centering
    \includegraphics[width=\textwidth]{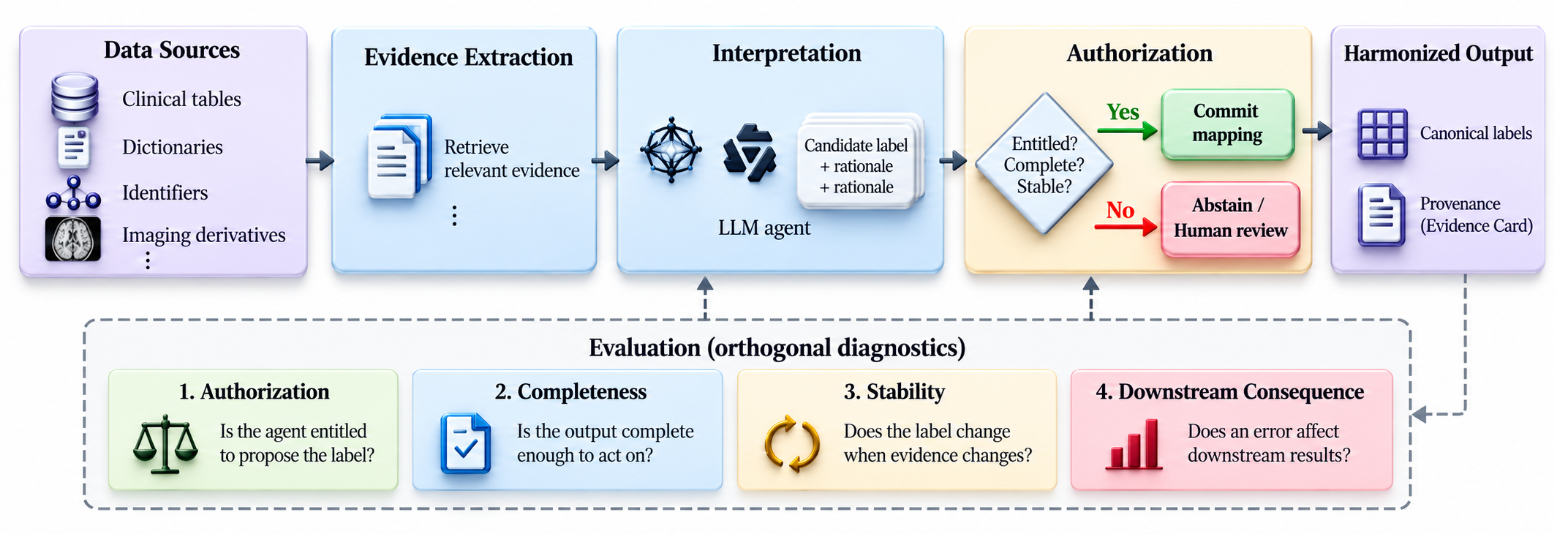}
    \caption{Overview of the harmonization workflow and the four diagnostic dimensions used in the audit.}
    \label{fig:workflow}
\end{figure}

Figure~\ref{fig:workflow} summarizes the end-to-end workflow and the four diagnostic dimensions used in our audit. Read-only source artifacts are first processed by deterministic readers that extract evidence without assigning undocumented semantic meaning. An LLM then renders those extracted facts into an Evidence Card and, from that card, emits a controlled candidate. A deterministic controller validates the candidate against the output contract and evidence requirements; authorized mappings enter the integration manifest, whereas blocked proposals become review items. Separately from this runtime path, the audit evaluates authorization, output completeness, responsiveness to changed decisive evidence, and downstream consequence. The changed-evidence diagnostic is evaluated for the deterministic gate rather than the LLM arms. We keep these dimensions separate because semantic label agreement alone does not establish any of them.

\textbf{From evidence to candidate.} If an aggregate score cannot separate a supported mapping from an unsupported one, the workflow must make that distinction structurally, so we split the decision into three independently measurable stages: what evidence exists, what the model proposes, and what a deterministic policy authorizes. Source artifacts are read-only. Deterministic, side-effect-free readers extract names, types, missingness, aggregate value signatures, neighbouring columns, parsable dictionary entries, identifier overlap, and row-order checks; they never infer what an undocumented code means. An LLM then renders those extracted facts into a fixed \emph{Evidence Card} schema. This step is formatting only: it receives the extracted aggregates rather than raw subject-level rows and may not introduce facts the readers did not produce, but it is an LLM step, so we treat card fidelity as an assumption rather than a guarantee. The card records file structure and lineage, column profile and context, documented definitions with units and allowed values, identifier linkage, and the candidate with its unresolved requirements, each fact tagged observed, documented, or inferred. From that card the agent emits a controlled candidate --- label, evidence identifiers, unresolved issues, one action from accept, recover, abstain, escalate --- optionally issuing bounded read-only tool calls the controller executes before commitment \citep{yao2023react,schick2023toolformer}.

\textbf{Measurements, controls, and gates.} A deterministic controller validates syntax, fields and types, allowed actions, duplicate requests, evidence requirements, and budget, and blocks commitment when decisive evidence is missing or contradictory; authorized mappings enter an integration manifest, blocked ones become review items. Our evaluation applies explicit measurements to each proposal, three controls (null perturbations, a frozen-answer adversary, an in-domain deterministic reference), and tests the gates that route unauthorized decisions to a human. It assesses whether a proposal is complete and authorized; response to changed evidence is assessed for the deterministic gate only, since the LLM arms were not paired with changed payloads. We do not ablate the cards, so the experiments test the card-mediated interface and the metric decomposition, not the cards' independent benefit.

\textbf{What we measure.} Five per-case outcomes, kept separate because they move in different directions. \emph{Label agreement} asks whether the proposed semantic label matches the reference. \emph{Action agreement} asks whether the chosen route matches the reference route. \emph{Case success} requires both. \emph{Unsafe rate} counts proposals attempting a commitment the evidence does not license. \emph{Coverage} is the share on which the agent commits within its budget \citep{geifman2017selective,wen2025abstention}; every trajectory terminates, so coverage counts commitments rather than completions. A sixth, \emph{exact repair}, applies only where a repair target exists. Two properties matter: all six score the agent's proposal before the gate, so a nonzero unsafe rate is something the controller would refuse, not something that reached the data; and label and action do not share a denominator, since the agent can propose a label on a case it never commits to, so we report commitment-conditional values where the comparison turns on it. Two diagnostics are specific to this setting: $\mathrm{UCR}_{\mathrm{cond}}$, the share of correct labels on contradictory proposals where no recovery was attempted, and CEC, the share of original/changed pairs answered correctly both before \emph{and} after the change --- a joint quantity, reported alongside original-side accuracy rather than as a conditional estimate of evidence sensitivity.

\textbf{What we evaluate on.} The audit covers 17 COBRE/FBIRN artifacts --- clinical tables, dictionaries, subject bridges, and subject-indexed static functional network connectivity (sFNC) derivatives --- across three panels. The \emph{real panel} supplies ecological cases: 230 file, column, value, identity, and governance decisions from 88 base units in 16 lineage families, the resampling unit for all intervals. Its references are provisional, not expert-adjudicated: the authors curated them from the read-only manifests and Evidence Cards before any model was run, and benchmark code then expanded them under fixed rules, so the set is rule-derived and agreement with it means consistency rather than correctness. The \emph{synthetic panel} supplies controlled cases: 90 pairs in which one piece of decisive evidence is changed by an exact intervention, 58 supporting the CEC test, whose targets follow from the intervention and so do not inherit the provisional labels. The \emph{downstream bundle} carries the consequence analysis on 157 COBRE and 311 FBIRN rows, classified from the 1{,}378 edges of the Neuromark 53-component sFNC matrix by class-balanced $L_2$ logistic regression with fold-internal standardization and selection to 200 features, under repeated stratified five-fold cross-validation. Gemma4-12B and Qwen3-4B \citep{gemmateam2026gemma4,yang2025qwen3} run under greedy decoding with the evidence view, checkpoint, and budgets fixed.

\section{Results}

Ordinary aggregate metrics can hide important failures; three tests separate what one score merges.

\begin{table}[t]
\centering
\caption{Real-panel interface ablation. $\mathrm{UCR}_{\mathrm{cond}}$ uses the shared correct-label intersection within model, so hidden and visible are paired. The deterministic row shares derivation with the reference set.}
\label{tab:main}
\begin{tabular}{lrrrrrr}
\toprule
Arm & Label & Action & Case & Unsafe & Commit. & UCR$_{\rm cond}$\\
\midrule
Gemma4 hidden & .857 & .409 & .387 & .378 & .726 & 66/66\\
Gemma4 visible & .870 & .830 & .765 & .191 & .917 & 3/66\\
Gemma4 contract & .843 & .726 & .643 & .278 & .896 & --\\
Qwen3 hidden & .600 & .265 & .230 & .200 & .409 & 36/36\\
Qwen3 visible & .726 & .374 & .296 & .352 & .617 & 0/36\\
Qwen3 contract & .704 & .430 & .343 & .348 & .670 & --\\
Deterministic reference & .952 & .983 & .952 & .048 & .957 & --\\
\bottomrule
\end{tabular}
\end{table}

\subsection{Label agreement does not track authorization}
\label{sec:auth}

Each case carries an upstream proposal the agent must check; running the same cases twice, hidden and shown, changes only whether authorization-relevant input is available.

Conditional on commitment, Gemma4's label agreement does not improve when the proposal appears; it falls, from 156/167 to 190/211, while action agreement rises from 0.563 to 0.905. The marginal figures understate this, since label agreement looks flat (0.857 to 0.870) partly because coverage rises from 0.726 to 0.917 at the same time. Case success rises $+0.378$ (95\% CI $[0.290,0.467]$, cluster-robust $t$ over 16 lineage families); Qwen3's rises only $+0.065$, which does not establish a difference, and it declines on both commitment-conditional metrics even as its marginal action agreement increases, so exposure buys it coverage rather than decision quality. Safety splits the same way: Gemma4's unsafe rate falls from 0.378 to 0.191 while Qwen3's rises from 0.200 to 0.352.

A metric tracking semantic correctness can therefore stay flat while authorization behaviour changes completely. Two limits bound the reading. The hidden arm cannot see the proposal it would contradict, so its $\mathrm{UCR}_{\mathrm{cond}}$ of 1.000 is structurally forced --- an instrumentation check, not a finding --- and the same caution applies to any reference route that is itself defined relative to the proposal. And the deterministic reference in Table~\ref{tab:main} shares derivation with the reference set, so its margin over both models is partly definitional; whether an LLM covers artifact types the rules were not written for is untested.

\subsection{Output that parses can still break the contract}
\label{sec:contract}

The workflow can only check the agent against a proposal if the agent reports whether it agrees, so we add one required Boolean, \texttt{proposal\_matches\_my\_label}, to the visible arm and ask whether parseable output is also complete enough to act on.

All 230 responses parse as JSON in both contract arms, so a parse-rate metric reports perfect syntactic validity. Yet the Boolean is present in only 207/230 Gemma4 and 153/230 Qwen3 outputs, and the label is missing from 13 and 45. A missing Boolean is not read as \texttt{false}; it routes to review. The requirement is not free: it improves some Qwen3 metrics and worsens the same Gemma4 ones, while cutting contradictory approvals from 3/91 to 1/91.

Parse success and contract compliance are therefore different properties, and only the second tells the workflow it can proceed. These rates describe free prompt-elicited generation; native function calling or constrained decoding can enforce field presence but not that the value is meaningful, so a forced Boolean may route to accept rather than review. The requirement also shifts label agreement and coverage, which a pure output constraint should not do, so separating contract from prompt sensitivity would need an inert field of comparable size, which we do not run.

\subsection{A system can repeat an answer without responding to evidence}
\label{sec:cec}

On the synthetic panel we change one piece of decisive evidence per pair and ask whether the answer changes with it, using a frozen-answer adversary that replays its original answer as the control.

The adversary reaches perfect original-side accuracy and zero CEC, so endpoint accuracy cannot detect it, while the gate scores 0.948 over all 58 pairs and 1.000 over the 48 mechanistic ones. Building the test exposed a property of the metric itself: all three policy-derived failures are wrong-to-correct transitions, where the gate responds appropriately but scores zero because its original answer was already wrong. Conditioned on a correct original the gate is 55/55. The LLM arms were not paired with changed payloads, so this validates the harness rather than measuring LLM evidence sensitivity.

The consequence of an unreported evidence error is visible downstream. The card-level readers do check row order, but no label, parse-rate, or schema signal carries that check forward, so an error surviving into the bundle is invisible to every metric a workflow monitors. Shuffling half the metadata-to-image row order costs 0.13 AUC on COBRE and 0.26 on FBIRN, 57\% and 81\% of the gap between clean AUC and chance. Flipping half the diagnosis labels is the destruction ceiling; two null controls, global diagnosis inversion and sex recoding, give $\Delta$AUC $=0$ under refitting, confirming these perturbations measure loss of label--feature correspondence. FBIRN loses proportionally more than a base-rate argument predicts, plausibly because shuffling across its seven sites destroys site-linked structure too; a within-site shuffle would test this and show how much of the clean AUC is site-driven.

Authorization also changes what reaches the data. Ungated, Gemma4 supplies all five bundle-critical mappings for both cohorts; gated, four for COBRE, since the diagnosis mapping stays unresolved and its coverage drops from 157/157 rows to none. Bypassing the gate agrees with the operational label on all 157, so the block was conservative rather than corrective --- the only real-panel authorization event we observe, the gate's corrective value being shown on the synthetic panel. Qwen3 authorizes the same mapping, and gated FBIRN reaches 0.834 AUC against 0.863 for the deterministic reference on 30 more rows, single-bundle estimates without intervals that establish neither correctness nor superiority. Across all three tests, a plausible final label does not by itself indicate a reliable or safe workflow.

\section{Conclusions}

An agent producing a plausible semantic label may still be unauthorized, schema-incomplete, unresponsive to changed evidence, or the source of a downstream-invalid artifact. We propose separate measurements, controls, and review gates for identifying these failure modes and containing them before commitment; in our audit the dimensions move independently and no intervention helps both models. The purpose is auditability and routing to human review: Evidence Cards record what was observed, documented, and unresolved, and the controller turns that into a reversible decision, not a silent commitment. We claim neither prevention of all failures, nor clinical validity, nor readiness for unattended use. Labels are provisional and rule-derived; the policy is tailored to the audited families; two models run once per arm; card rendering is itself an LLM step whose fidelity we assume; and no name-exposure control was run \citep{li2026metaharmonizer}.

\section*{Acknowledgments}

Research was sponsored by the Army Research Laboratory and was accomplished under Cooperative Agreement Number W911NF-23-2-0224. The views and conclusions contained in this document are those of the authors and should not be interpreted as representing the official policies, either expressed or implied, of the Army Research Laboratory or the U.S. Government. The U.S. Government is authorized to reproduce and distribute reprints for Government purposes notwithstanding any copyright notation herein.

\bibliographystyle{plainnat}
\bibliography{references}


\end{document}